\documentclass{elsart}

\usepackage{natbib,amsmath,array,multicol}

\newcommand{\CollagenTM}{\textsc{Collagen}$^{\text{\tiny{\textsf{(TM)}}}}$}

\newcommand{\Action}[1]{$<$\textsl{#1}$>$}
\newenvironment{InTable}{\setlength{\extrarowheight}{-5pt}\begin{tabular}[t]{l}}{\end{tabular}}
\newenvironment{InTableII}{\setlength{\extrarowheight}{-5pt}\begin{tabular}[t]{ll}}{\end{tabular}}

\newif\ifpdf
\ifx\pdfoutput\undefined
  \pdffalse
\else
  \pdfoutput=1
  \pdftrue
\fi

\ifpdf
  \usepackage[pdftex,final]{graphicx}
\else
  \usepackage[dvips,final]{graphicx}
\fi

\usepackage{amssymb}

\begin{document}

\begin{frontmatter}



\title{Explorations in Engagement\\ for Humans and Robots}


 \author[merl]{Candace L. Sidner\corauthref{cor1}}
 \ead{sidner@merl.com}
 \ead[url]{http://www.merl.com/people/sidner}
 \corauth[cor1]{Corresponding author}
 \author[merl]{Christopher Lee}
 \author[mit]{Cory Kidd}
 \author[merl]{Neal Lesh}
 \author[merl]{Charles Rich}
\address[merl]{Mitsubishi Electric Research Laboratories\\
201 Broadway,\\
Cambridge, MA 02139 USA}
\address[mit]{The Media Laboratory\\
Massachusetts Institute of Technology\\
77 Massachusetts Avenue\\
Cambridge, MA 02139 USA}

\begin{abstract}
  This paper explores the concept of engagement, the process by which
  individuals in an interaction start, maintain and end their
  perceived connection to one another.  The paper reports on one
  aspect of engagement among human interactors---the effect of
  tracking faces during an interaction.  It also describes the
  architecture of a robot that can participate in conversational,
  collaborative interactions with engagement gestures.  Finally, the
  paper reports on findings of experiments with human participants who
  interacted with a robot when it either performed or did not perform
  engagement gestures.  Results of the human-robot studies indicate
  that people become engaged with robots: they direct their attention
  to the robot more often in interactions where engagement gestures
  are present, and they find interactions more appropriate when
  engagement gestures are present than when they are not.

\end{abstract}


\end{frontmatter}

\section{Introduction}

When individuals interact with one another face-to-face, they use
gestures and conversation to begin their interaction, to maintain and
accomplish things during the interaction, and to end the interaction.
Engagement is the process by which interactors start, maintain and end
their perceived connection to each other during an interaction.  It
combines verbal communication and non-verbal behaviors, all of which
support the perception of connectedness between interactors.  While
the verbal channel provides detailed and rich semantic information as
well as social connection, the non-verbal channel can be used to
provide information about what has been understood so far, what the
interactors are each (or together) attending to, evidence of their
waning connectedness, and evidence of their desire to disengage.

Evidence for the significance of engagement becomes apparent in
situations where engagement behaviors conflict, such as when the
dialogue behavior indicates that the interactors are engaged (via turn
taking, conveying intentions and the like), but when one or more of
the interactors looks away for long periods to free space or objects
that have nothing to do with the dialogue.  This paper explores the
idea that engagement is as central to human-robot interaction as it is
for human-human interaction\@.\footnote{The use of the term
  ``engagement'' was inspired by a talk given by Alan Bierman at User
  Modelling 1999.  Bierman (personal communication, 2002) said ``The
  point is that when people talk, they maintain conscientious
  psychological connection with each other and each will not let the
  other person go.  When one is finished speaking, there is an
  acceptable pause and then the other \emph{must} return something.  We have
  this set of unspoken rules that we all know unconsciously but we all
  use in every interaction. If there is an unacceptable pause, an
  unacceptable gaze into space, an unacceptable gesture, the
  cooperating person will change strategy and try to re-establish
  contact.  Machines do none of the above, and it will be a whole
  research area when people get around to working on it.''}

Engagement is not well understood in the human-human context, in
part because it has not been identified as a basic behavior.  Instead,
behaviors such as looking and gaze, turn taking and other
conversational matters have been studied separately, but only in the
sociological and psychological communities as part of general
communication studies.  In artificial intelligence, much of the focus
has been on language understanding and production, rather than gestures
or on the fundamental problems of how to get started and
stay connected, and the role of gesture in connecting.  Only with the
advent of embodied conversational (screen-based) agents and better
vision technology have issues about gesture begun to come forward (see
\cite{traum02:_embod_agent} and \cite{nakano03:_face_groun} for
examples of screen-based embodied conversational agents where these
issues are relevant).

The methodology applied in this work has been to study human-human
interaction and then to apply the results to human-robot interaction,
with a focus on hosting activities.  Hosting activities are a class of
collaborative activity in which an agent provides guidance in the form
of information, entertainment, education or other services in the
user's environment.  The agent may also request that the user
undertake actions to support its fulfillment of those services.
Hosting is an example of what is often called ``situated''or
``embedded'' activities, because it depends on the surrounding
environment as well as the participants involved.  We model hosting
activities using the collaboration and conversation models of
\cite{grosz86:_atten}, \cite{grosz96:_collab_plans}, and
\cite{lochbaum98:_collab_plann}.  Collaboration is distinguished from
those interactions in which the agents cooperate but do not share goals.

In this work we define interaction as an encounter between two or more
individuals during which at least one of the individuals has a purpose
for encountering the others.  Interactions often include conversation
although it is possible to have an interaction where nothing is
communicated verbally.  Collaborative interactions are those in which
the participating individuals come to have shared goals and intend to
carry out activities to attain these shared goals.  This work is
directed at interactions between only two individuals.

Our hypothesis for this work concerned the effects of engagement
gestures during collaborative interactions.  In particular, we expect
that a robot using appropriate looking gestures and one that had no
such gestures would differentially affect how the human judged the
interaction experience.  We further predicted that the human would
respond with corresponding looking gestures whenever the robot looked
at and away from the human partner in appropriate ways.  The first
part of this paper investigates the nature of looking gestures in
human-human interactions.  The paper then explains how we built a
robot to approximate the human behavior for engagement in
conversation.  Finally, the paper reports on an experiment wherein a
human partner either interacts with a robot with looking gestures or
one without them.  A part of that experiment involved determining
measures to use to evaluate the behavior of the human interactor.

\section{Human-human engagement: results of video analysis}

This section presents our work on human-human engagement.  First we
review the findings of previous research that offer insight into the
purpose of undertaking the current work.

Head gestures (head movement and eye movement) have been of interest
to social scientists studying human interaction since the 1960s.
\cite{argyle76:_mutual_gaze} documented the function of gaze as an
overall social signal, to attend to arousing stimulus, and to express
interpersonal attitudes, and as part of controlling the
synchronization of speech. They also noted that failure to attend to
another person via gaze is evidence of lack of interest and attention.
Other researchers have offered evidence of the role of gaze in
coordinating talk between speakers and hearers, in particular, how
gestures direct gaze to the face and why gestures might direct it away
from the face
(\cite{kendon67:_gaze_direction,duncan72:_signal_rules,heath86:_body_movem,goodwin86:_gestur}
among others).  Kendon's observations (1967) that the participant
taking over the turn in a conversation tends to gaze away from the
previous speaker has been widely cited in the natural language
dialogue community.  Interestingly, Kendon thought this behavior might
be due to the processing load of organizing what was about to be said,
rather than a way to signal that the new speaker was undertaking to speak.
More recent research argues that the information structure of the turn
taker's utterances governs the gaze away from the other participants
(\cite{cassell99:_turn_takin}).

Other work has focused on head movement alone
(\cite{kendon70:_movem,mcclave00:_linguis}) and its role in
conversation.  Kendon looked at head movements in turn taking and how
they were used to signal change of turn, while McClave provided a
large collection of observations of head movement that details the use
of head shakes and sweeps for inclusion, intensification or
uncertainty about phrases in utterances, change of head position to
provide direct quotes, to provide images of characters and to place
characters in physical space during speaking, and head nods as
backchannels and as encouragement for listener
response\@.\footnote{\cite{yngve70} first observed the use of nods as
  backchannels, which are gestures and phrases such as ``uh-huh,
  mm-hm, yeh, yes'' that hearers offer during conversation.  There is
  disagreement about whether the backchannel is used by the hearer to
  take a turn or to avoiding doing so.}

While these previous works provide important insights as well as
methodologies for how to observe people in conversation, they did not
intend to explore the qualitative nature of head movement, nor did
they attempt to provide general categories into which such
behaviors could be placed.  The research reported in this paper has
been undertaken with the belief that regularities of behavior in head
movement can be observed and understood.  This work does not consider
gaze because it has been studied more recently in AI models for turn
taking (\cite{thorisson97:_gandalf,cassell99:_turn_takin}) and because
the operation of gaze as a whole for an individual speaker and for an
individual listener is still an area in need of much research.  Nor is
this work an attempt to add to the current theories about looking and
turn taking.  Rather this work is focused on attending to the face of
the speaker, and harks back to Argyle and Cook's (1976) ideas about
looking (in their studies, just gazing) as evidence of interest.  Of
most relevance to gaze, looking and turn taking is Nakano et al's
recent work on grounding, which reports on the use of the hearer's gaze
and the lack of negative feedback to determine whether the speaker's
turn has been grounded by the hearer.  As will be clear in the next
section, our observations of looking behavior complement the empirical
findings of that work.

The robotic interaction research reported in this paper was inspired by work
on embodied conversation agents (ECAs).  The Steve system, which
provided users a means to interact with the ECA Steve through
head-mounted glasses and associated sensors, calculated the
user's field of view to determine which objects were in view, and used
that information to generate references in utterances
(\cite{rickel99:_animat_agent}).  Other researchers (notably,
\cite{rea2000,cassell00:_embod_conver_agent,johnson00:_animat_pedag_agent},
\cite{gratch02:_virtual_humans}) have developed ECAs that produce
gestures in conversation, including facial gestures, hand gestures and
body movements.  However, they have not tried to incorporate
recognition as well as production of these gestures, nor have they
focused on the use of these behaviors to maintain engagement in
conversation.

One might also consider whether people necessarily respond to robots
in the same way as they do to screen-based agents.  While this topic
requires much further analysis, work by \cite{kidd03:_sociab_robot}
indicates that people collaborate differently with a telepresent robot
than with a physically present robot.  In that study, the same robot
interacted with all participants, with the only difference being that
for some participants the robot was present only by video link (i.e.,
it appeared on screen to interact with a person).  Participants found
the physically present robot more altruistic, more persuasive, more
trustworthy, and providing better quality of information.

For the work presented here, we videotaped interactions of two people in a
hosting situation, and transcribed portions of the video for all the
utterances and some of the gestures (head, body position, body
addressing) that occurred.  We then considered one behavior in detail,
namely mutual face tracking of the participants, as evidence of their
focus of interest and engagement in the interaction.  The purpose of
the study was to determine how well the visitor (V) in the hosting
situation tracked the head motion of the host (H), and to characterize
the instances when V failed to track H\@.\footnote{We say that V
  ``tracks H's changes in looking'' if: when H looks at V, then V
  looks back at H; and when H looks elsewhere, V looks toward the same
  part of the environment as H looked.}
While it is not possible to draw conclusions about all human behavior
from a single pair interaction, even a single pair provides an
important insight into the kinds of behavior that can occur.

In this study we assumed that the listener would track the speaker
almost all the time, in order to convey engagement and use non-verbal
as well as verbal information for understanding.   In our study the
visitor is the listener in more than 90\% of the interaction (which is
not the normal case in conversations)\@.\footnote{The visitor says
  only 15 utterances other than 43 backchannels (for example, ok,
  ah-hah, yes, and wow) during 5 minutes and 14 seconds of dialogue.
  Even the visitor's utterances are brief, for example,
  \emph{absolutely, that's very stylish, it's not a problem}.}

\begin{table}
  \centering
  \setlength{\extrarowheight}{-2pt}
  \begin{tabular}[c]{|l|c|c|c|}\hline
    &  & \multicolumn{2}{c|}{\emph{Percentage of:}} \\\cline{3-4}
    & Count & Tracking failures & Total host looks \\\hline\hline
    Quick looks   & 11 & 30\% & 13\% \\
    Nods          & 14 & 38\% & 17\% \\
    Uncategorized & 12 & 32\% & 15\% \\\hline
  \end{tabular}
  \caption{Failures of a visitor (V) to track changes in host's (H) looking
    during a conversation.}
  \label{tab:tracking-failures}
\end{table}

To summarize, there are 82 instances where the (male) host (H) changed
his head position, as an indication of changes in looking, during a
five minute conversational exchange with the (female) visitor (V).
Seven additional changes in looking were not counted because it was
not clear to where the host turned.  Of his 82 counted changes in
looking, V tracks 45 of them (55\%).  The remaining failures to track
looks (37, or 45\% of all looks) can be subclassed into 3 groups:
\emph{quick looks} (11), \emph{nods} (14), and \emph{uncategorized
  failures} (12), as shown in Table~\ref{tab:tracking-failures}.  The
quick look cases are those for which V fails to track a look that
lasts for less than a second.  The nod cases are those for which V
nods (e.g., as an acknowledgement of what is being said) rather than
tracking H's look.

The quick look cases happen when V fails to notice H's look due to
some other activity, or because the look occurs in mid-utterance and
does not seem to otherwise affect H's utterance.  In only one instance
does H pause intonationally and look at V\@.  One would expect an
acknowledgement of some kind from V here, even if she doesn't track
H's look, as is the case with nod failures.  However, H proceeds even
without the expected feedback.

The nod cases can be explained because they occur when H looks at V
even though V is looking at something else.  In all these instances, H
closes an intonation phase, either during his look or a few words
after, to which V nods and often articulates with ``Mm-hm,'' ``Wow''
or other phrases to indicate that she is following her conversational
partner.  In grounding terms (\cite{clark96:_using_languag}), H is
attempting to ascertain by looking at V that she is following his
utterances and actions.  When V cannot look, she provides feedback by
nods and comments.  She is able to do this because of linguistic (that
is, prosodic) information from H indicating that her contribution is
called for.

Of the uncategorized failures, the majority (8 instances) occur when V
has other actions or goals to undertake.  In addition, all of the
uncategorized failures are longer in duration than quick looks (2
seconds or more).  For example, V may be finishing a nod and not be
able to track H while she's nodding.  Of the remaining three tracking
failures, each occurs for seemingly good reasons to video
observers, but the host and visitor may or may not have been aware of
these reasons at the time of occurrence.  For example, one failure
occurs at the start of the hosting interaction when V is looking at
the new (to her) object that H displays and hence does not track H when he
looks up at her.

Experience from this data has resulted in the \emph{principle of
  conversational tracking}: a participant in a collaborative
conversation tracks the other participant's face during the
conversation in balance with the requirement to look away in order to:
(1) participate in actions relevant to the collaboration, or (2)
multi-task with activities unrelated to the current collaboration,
such as scanning the surrounding environment for interest or danger,
avoiding collisions, or performing personal activities.

\section{Applying the results to robot behavior}

\begin{figure}
  \centering
  \includegraphics[width=4in]{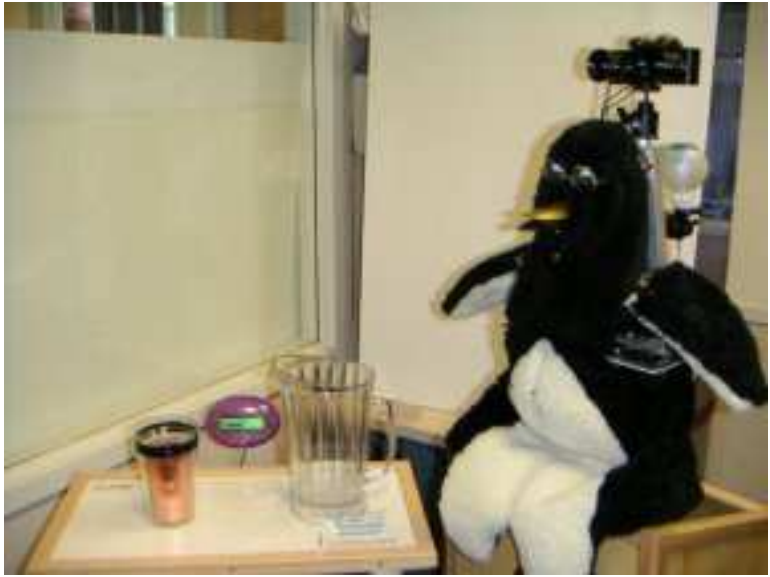}
  \caption{Mel, the penguin robot with the IGlassware table}
  \label{fig:melvin-pic}
\end{figure}
The above results and the principle of conversational tracking have
been put to use in robot studies via two different gesture strategies,
one for behavior produced by the robot and one for interpreting user
behavior.  Our robot, named Mel, is designed to resemble a penguin
wearing glasses (Figure~\ref{fig:melvin-pic}), and is described in
more detail in Section~\ref{sec:architecture}.  The robot's default
behavior during a conversation is to attend to the user's face, i.e.,
to keep its head oriented toward the user's face.  However, when
called upon to look at objects in the environment during its
conversational turn, the robot turns its head toward objects (either
to point or indicate that the object is being reintroduced to user
attention).  Because the robot is not mobile and cannot see other
activities going on around it, the robot does not scan the
environment.  Thus the non-task oriented lookaways observed in our
studies of a human speaker are not replicated in these strategies with
the robot.

A portion of the robot's verbal behavior is coordinated with gestures
as well.  The robot converses about the task and obeys a model of turn
taking in conversation.  The robot always returns to face the user
when it finishes its conversational turn, even if it had been directed
elsewhere.  It also awaits verbal responses not only to questions, but
to statements and requests, to confirm user understanding before it
continues the dialogue. This behavior parallels that of the human
speaker in our studies.  The robot's collaboration and conversation
abilities are based on the use of a tool for collaborative
conversation (\cite{rich98:_collag,rich01:_collag}).  An example
conversation for a hosting activity is discussed in
Section~\ref{sec:architecture}.

In interpreting human behavior, the robot does not adhere to the
expectation that the user will look at the robot most of the time.
Instead it expects that the user will look around at whatever the user
chooses.  This expectation results from the intuition that users might
not view the robot as a typical conversational partner.  Only when the
robot expects the user to view certain objects does it respond if the
user does not do so. In particular, the robot uses verbal statements
and looking gestures to direct the user's attention to the object.
Furthermore, just as the human-human data indicates, the robot
interprets head nods as an indication of grounding\@.\footnote{We view
  grounding as a backward looking engagement behavior, one that
  solidifies what is understood up to the present utterance in the
  interaction.  Forward looking engagement tells the participants that
  they continue to be connected and aware in the interaction.}  Our
models treat recognition of user head nodding as a probabilistic
classification of sensed motion data, and the interpretation of each
nod depends on the dialogue context where it occurs.  Only head nods
that occur when or just before the robot awaits a response to a
statement or request (a typical grounding point) are interpreted as
acknowledgement of understanding.

The robot does not require the user to look at it when the user takes
a conversational turn (as is prescribed by \cite{sacks74}).  However,
as we discuss later, such behavior is typical in a majority of the
user interactions.  The robot \emph{does} expect that the user will
take a turn when the robot signals its end of turn in the
conversation.  The robot interprets the failure to do so as an
indication of disengagement, to which it responds by asking whether
the user wishes to end the interaction.  This strategy is not based on
our human-human studies, since we saw no instances where failure to
take up the turn occurred.

The robot also has its own strategies for initiating and terminating
engagement, which are not based on our human-human studies. The robot
searches out a face while offering greetings and then initiates
engagement once it has some certainty (either through user speech or
close proximity) that the user wants to engage (see the discussion in
Section~\ref{sec:architecture} for details on how this is
accomplished).  Disengagement occurs by offering to end the
interaction, followed by standard (American) good-bye rituals
(\cite{schegeloff73:_closings}), including the robot's looking away
from the user at the close.

\section{Architectures to support human-robot engagement,
  collaboration and conversation.}
\label{sec:architecture}

Successful interaction between the human and robot requires the robot
express its own engagement, and to interpret the human's engagement
behavior.  This section reports on an architecture and its components
to support engagement in collaborative interactions.

\begin{figure}
  \centering
  \includegraphics[width=4in]{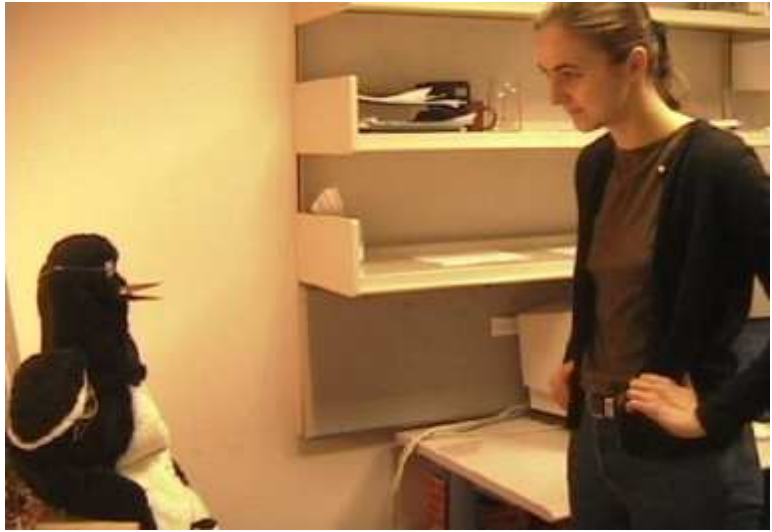}
  \caption{Mel demonstrates IGlassware to a visitor.}
  \label{fig:demonstration}
\end{figure}
\begin{figure}
\newcommand{\M}[1]{\item[\small\textsc{Mel}:]\small#1}
\newcommand{\U}[1]{\item[\small\textsc{User}:]\small\textsl{#1}}
\newcommand{\A}[1]{\item[]\small\textsl{#1}}
\setlength{\extrarowheight}{-5pt}
\setlength{\tabcolsep}{2pt}
\begin{multicols}{2}
\begin{list}{}{%
\setlength{\labelwidth}{1em}
\setlength{\leftmargin}{0.5em}
\setlength{\itemindent}{0pt}
\setlength{\rightmargin}{0pt}
\setlength{\itemsep}{1pt}}
\A{Robot moves its head back and forth until it locates a talking face.}
\M{Hi, I'm Mel a robotic penguin.}
\U{Hi.}
\A{After this point, the robot tracks the user's face as the user
  moves around or turns his or her head.}
\M{What's your name?}
\U{Sam.}
\M{Sam, I'd like to show you a demo. OK?}
\U{No.}
\M{But it's really interesting.  Come on.  Try it!}
\U{Ok.}
\M{It would be really nice to know what kind of progress people are
  making in their dining. So Paul Dietz created a new product called
  IGlassware for this.  By the way, Sam, do you know Paul?}
\U{No.}
\M{Well, Paul is a member of MERL\@.  This demo is one of his creations.}
\U{Ok.}
\M{IGlassware stands for Instrumented glassware!  Sam, right there
  \Action{robot turns head to table} is the IGlassware
  cup. \Action{robot turns head to user}}
\A{Some users do not turn to the cup, so the robot says:}
\M{The cup is here to my \Action{robot turns head to table} right
  \Action{robot turns head to user}.}
\A{User turns toward the cup.}
\M{And near it, is the table readout.  The word re-fill on the readout
  means that it is empty.}
\U{Ok.}
\M{All right, now we'll see how to use the cup.}
\U{All right.}
\M{First you should pour enough water from the pitcher into the glass
  to fill it up.  Then make sure the glass is on the IGlassware
  table.}
\U{Ok. \Action{User pours water}}
\M{Good, notice, the bar on the readout \Action{robot turns head to
    readout display on table} shows that the glass is full
  \Action{robot turns back to user}.}
\U{Right.}
\M{Ok, pour the water back into the pitcher.}
\U{Ok. \Action{User pours water}}
\M{I'm waiting for a reading from the table for the glass.}
\M{Good. See, it registers needing a re-fill!}
\U{Yes.}
\M{Would you like me to explain how this works?}
\U{Sure.}
\M{The copper in the glass \Action{robot turns head to readout
    display} transmits to the readout display by inductance with the
  surface of the table. The readout then displays the information
  coming to the table \Action{Robot turns head back to user}.}
\U{Cool.}
\M{Would you like to know how this technology might be used in restaurants?}
\U{Yes.}
\M{The glass tells the restaurant when the customer needs a refill.
  In restaurants drinks mean profit, so this gadget makes it easier
  for restaurants to sell more drinks and make more money.}
\U{Ok.}
\M{Well, Sam, that's about all. Go see Paul Dietz, for more about
  IGlassware. So long!}
\U{Good-bye.}
\end{list}
\end{multicols}
  \caption{Example conversation between Mel and a human user.}
  \label{fig:conversation}
\end{figure}

The robot's interaction abilities have been developed and tested using
a target task wherein the robot, acting as host, collaboratively demonstrates a
hardware invention, IGlassware (\cite{dietz02:_iglass}), to a human visitor
(Figure~\ref{fig:demonstration}).  The robot is designed to resemble a
penguin wearing glasses, and is stationary.  Because the robot has only wings but no hands, it relies on
the human to perform the physical manipulations necessary
for the demonstration.  The human thus must agree to collaborate for
the demo to succeed.  A typical interaction lasts about 3.5 minutes
and an example is shown in
Figure~\ref{fig:conversation}.  Robot beat gestures\@,\footnote{%
  Beat gestures are hand or occasionally head movements that are
  hypothesized to occur to mark new information in an utterance
  (\cite{cassell00:_nudge,cassell01:_beat}).}  head nods, and generic
human gestures are not included in the figure.  If the human does not
agree to participate in the demo, the robot engages in brief, basic
social ``chit-chat'' before closing the conversation.  How the user
responds to the robot's looks at the table are discussed in Section
\ref{userstudies}.

The robot's hardware consists of:
\begin{itemize}
\item 7 servos (two 2 DOF shoulders, 2 DOF neck, 1 DOF beak)
\item Stereo camera (6 DOF head tracking
       software of \cite{morency03:_pose_estim,viola01:_rapid_detec})
\item Stereo microphones (with speech detection and direction-location
  software)
\item Far-distance microphone for speech recognition
\item 3 computers: one for sensor fusion and robot motion,
  one for vision (6 DOF head tracking and head-gesture
  recognition), one for dialogue (speech recognition, dialogue modeling,
  speech generation and synthesis).
\end{itemize}

Our current robot is able to: 
\begin{itemize}
\item
  Initiate an interaction by visually locating a potential human
  interlocutor and generating appropriate greeting behaviors,
\item 
  Maintain engagement by tracking the user's moving face and
  judging the user's engagement based on head position (to the robot,
  to objects necessary for the collaboration),
\item 
  Reformulate a request upon failure of the user to respond to robot
  pointing,
\item 
  Point and look at objects in the environment,
\item 
  Interpret nods as backchannels and agreements in conversation \cite{kapoor01:_nod_detector,morency05:_nodding}, and
\item 
  Understand limited spoken utterances and produce rich verbal spoken
  conversation, for demonstration of IGlassware, and social ``chit-chat,''
\item 
  Accept appropriate spoken responses from the user and make
  additional choices based on user comments,
\item 
  Disengage by verbal interaction and closing comments, and simple
  gestures, 
\item 
  Interpret user desire to disengage (through gesture and speech
  evidence).
\end{itemize}
Verbal and non-verbal behavior are integrated and occur fully autonomously.

\begin{figure}
  \centering
  \includegraphics[width=4in]{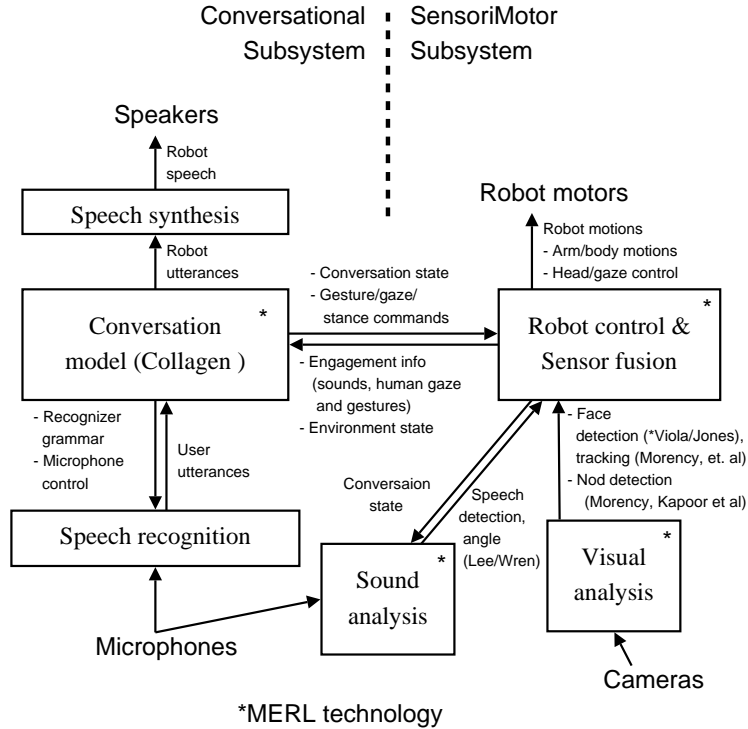}
  \caption{Robot software architecture}
  \label{fig:architecture}
\end{figure}

The robot's software architecture consists of distinct sensorimotor
and conversational subsystems.  The conversational subsystem is based
on the \CollagenTM{} collaboration and conversation model (see
\cite{rich98:_collag,rich01:_collag}), but enhanced to make use of
strategies for engagement.  The sensorimotor subsystem is a custom,
dynamic, task-based blackboard robot architecture.  It performs data
fusion of sound and visual information for tracking human
interlocutors in a manner similar to other systems such as
\cite{okuno03:_human_robot}, but its connection to the conversational
subsystem is unique.  The communication between these two subsystems
is vital for managing engagement in collaborative interactions with a
human.

\subsection{The Conversational Subsystem of the Robot}

For the robot's collaboration and conversation model, the special
tutoring capabilities of \CollagenTM{} were utilized.  In \CollagenTM{} a
task, such as demonstrating IGlassware, is specified by a hierarchical
library of ``recipes'', which describe the actions that the user and
agent will perform to achieve certain goals.  For tutoring, the
recipes include an optional prologue and epilogue for each action, to
allow for the behavior of tutors in which they often describe the act
being learned (the prologue), demonstrate how to do it, and then recap
the experience in some way (the epilogue).

At the heart of the IGlassware demonstration is a simple recipe for
pouring water from a pitcher into a cup, and then pouring the water from the cup back
into the pitcher.  These are the physical actions the robot ``teaches.''  The rest of the demonstration is comprised of explanations
about what the user will see, uses of the IGlassware table, and so on.  The interaction as a whole is described by a
recipe consisting of a greeting, the demonstration and a closing.  The demonstration is
an optional step, and if not undertaken, can be followed by an
optional step for having a short chat about
visiting the MERL lab. Providing these and other more detailed recipes to \CollagenTM{} makes it possible for the robot to interpret and participate in
the entire conversation using the built-in functions provided by
\CollagenTM{}\@.

Figure~\ref{fig:segmentedhistory} provides a representation, called a \emph{segmented interaction history} which \CollagenTM{} automatically incrementally computes during the robot interaction.  The indentation in Figure~\ref{fig:segmentedhistory} reflects the hierarchical (tree) structure of the underlying recipe library.  The terminal nodes of the tree are the utterances and actions of the human and the robot, as shown in Figure~\ref{fig:demonstration}. The non-terminal nodes of the tree (indicated by square brackets) correspond to the goals and subgoals of the task model.  For example, the three lines in bold denote the three first level subgoals of the top level goal in the recipe library.  Many parts of the segmented interaction history have been suppressed in Figure~\ref{fig:segmentedhistory} to save space.

\begin{figure}
\small
\newcommand{\I}[2]{\hbox{}\hspace{#1em}{#2}\\}
\I{0}{Got face.}
\I{0}{[Done interacting about IGlassware.]}
\I{1}{1 \textbf{[Done greeting.]}}
\I{1}{2 \textbf{[Done demonstrating IGlassware.]}}
\I{2}{Mel says ``Sam, I'd like to show you a demo, Ok?''}
\I{2}{User says ``No.'' (\ldots)}
\I{2}{[Done providing prologue to demonstrating IGlassware.]}
\I{3}{[Done discussing conditions of demonstrating IGlassware.]}
\I{4}{[Done Mel introducing IGlassware.]}
\I{5}{Mel says ``It would be really nice to know\ldots.''}
\I{5}{User says ``Ok.''}
\I{4}{[Done discussing Paul.]}
\I{4}{[Done user looking at cup by showing.]}
\I{5}{[Done showing how to look at cup.]}
\I{6}{[Done user looking at cup.]}
\I{7}{Mel says ``Sam, right there is the IGlassware cup.''}
\I{7}{Mel says ``The cup is here to my right.''}
\I{7}{User looks at cup.}
\I{6}{[Done Mel providing epilogue to showing how to look at cup.]}
\I{4}{[Done user looking at readout by showing.]}
\I{2}{[Done filling and emptying the glass.]}
\I{3}{[Done user pouring water into the cup by showing.]}
\I{4}{[Done showing how to pour water into the cup.]}
\I{5}{[Done user pouring water into the cup.]}
\I{3}{[Done user pouring water back into the pitcher by showing.]}
\I{4}{[Done showing how to pour water back into the pitcher.]}
\I{5}{[Done user pouring water back into the pitcher.]}
\I{6}{Mel says ``Ok. pour the water back into the pitcher.''}
\I{6}{User pours water back into the pitcher.}
\I{5}{[Done Mel providing epilogue to showing how to pour water back.]}
\I{6}{Mel says ``Good.''}
\I{6}{[Done discussing conditions of showing how to pour water back.]}
\I{7}{Mel says ``See, it registers needing a re-fill!''}
\I{7}{User says ``Yes.''}
\I{2}{[Done providing epilogue to demonstrating IGlassware.]}
\I{3}{[Done discussing conditions of demonstrating IGlassware.]}
\I{4}{[Done explaining how IGlassware works.]}
\I{4}{[Done explaining why the cup is useful.]}
\I{1}{3 \textbf{[Done closing by normal closing.]}}
\I{2}{Mel says ``Well, Sam, that's about all\ldots.''}
  \caption{Segmented Interaction History for Figure~{fig:demonstration}}
  \label{fig:segmentedhistory}
\end{figure}

The robot's language generation is achieved in two ways.  First,
\CollagenTM{} automatically produces a semantic representation of what
to say, which is appropriate to the current conversational and task
context.  For example, \CollagenTM{} automatically decides near the
beginning of the interaction to generate an utterance whose semantics
is a query for the value of an unknown parameter of a recipe, in this
case, the parameter corresponding to the user's name.  \CollagenTM{}'s
default realization for this type of utterance is ``what is the
\Action{parameter}?'' as in ``what is the user name?''  This default
is hardly a natural way to ask a person for their name.  To remedy
this problem, this default can be overriden by another part of the
generation algorithm in \CollagenTM{}.  It applies optional hand-built
application-specific templates.  In this example, it causes ``what is
your name?'' to be generated.  In addition, the robot's beat movements
and head turns are also hand-built to occur in concert with the
utterances.  Tracking the visitor's face and nodding at the user are
not hand crafted and occur automatically in the sensorimotor system.

Engagement behavior is integrated in \CollagenTM{} in two ways.  First,
engagement is a pervasive behavior rather than part of achieving any
particular goal, decisions about engagement (beginning it, determining
whether it is succeeding or failing, and when to end it) are handled
in \CollagenTM{}'s ``agent'' module.  The robot's \CollagenTM{} agent was
extended to provide additional decision-making rules for when a face
is found (so that greetings can occur), to determine when to abort the
demo, how to interpret looks away on the part of the user, and the
expectations that the user will look at specific objects during the
demo.

Second, engagement rules can introduce new goals into \CollagenTM{}'s
collaborative behavior.  For example, if the engagement rules
(mentioned previously) decide that the user is disengaging, a new goal
may be introduced to re-engage.  \CollagenTM{} will then choose among
its recipes to achieve the goal of re-engagement.  Thus the full
problem solving power of the task-oriented part of \CollagenTM{} is
brought to bear on goals which are introduced by the engagement layer.

\subsection{Interactions between the sensorimotor and conversational subsystems}
Interactions between the sensorimotor and conversational subsystems
flow in two directions.  Information about user manipulations and
gestures must be communicated in summary form as discrete events from
the sensorimotor to the conversational subsystem, so that the
conversational side can accurately model the collaboration and
engagement.  The conversational subsystem uses this sensory
information to determine whether the user is continuing to engage with
the robot, has responded to (indirect) requests to look at objects in
the environment, has nodded at the robot (which must be interpreted in
light of the current conversation state as either a backchannel, an
agreement, or as superfluous), is looking elsewhere in the scene, or
is no longer visible (a signal of possible disengagement).

In the other direction, high-level decisions and dialogue state must
be communicated from the conversational to the sensorimotor subsystem,
so that the robot can gesture appropriately during robot and user
utterances, and so that sensor fusion can appropriately interpret user
gestures and manipulations.  For example, the conversational subsystem
tells the sensorimotor subsystem when the robot is speaking and when
it expects the human to speak, so that the robot will look at the
human during the human's turn.  The conversational subsystem also
indicates the points during robot utterances when the robot should
perform a given beat gesture (\cite{cassell01:_beat}) in synchrony
with new information in the utterance, or when it should look at (only
by head position, not eye movements) or point to objects (with its
wing) in the environment in coordination with spoken output.  For
example, the sensorimotor subsystem knows that a \textsc{GlanceAt}
command from the conversational subsystem temporarily overrides any
default face tracking behavior when the robot is speaking.  However,
normal face tracking goes on in parallel with beat gestures (since
beat gestures in the robot are only done with the robot's limbs).

Our robot cannot recognize or locate objects in the environment.  In
early versions of the IGlassware demonstration experiments, we used
special markers on the cup so that the robot could find it in the
environment.  However, when the user manipulated the cup, the robot
was not able to track the cup quickly enough, so we omitted this type
of knowledge in more recent versions of the demo.  The robot learns
about how much water is in the glass, not from visual recognition, but
through wireless data that IGlassware sends to it from the table.

In many circumstances, information about the dialogue state must be
communicated from the conversational to the sensorimotor subsystem in
order for the sensorimotor subsystem to properly inform the
conversational subsystem about the environment state and any
significant human actions or gestures.  For example, the sensorimotor
subsystem only tries to detect the presence of human speech when the
conversational subsystem expects human speech, that is, when the robot
has a conversational partner and is itself not speaking.  Similarly,
the conversational subsystem tells the sensorimotor subsystem when it
expects, based on the current purpose as specified in its dialogue
model, that the human will look at a given object in the environment.
The sensorimotor subsystem can then send an appropriate semantic event
to the conversational subsystem when the human is observed to move
his/her head appropriately. For example, if the cup and readout are in
approximately the same place, a user glance in that direction will be
translated as \textsc{LookAt(human,cup)} if the dialogue context
expects the user to look at the cup (e.g., when the robot says ``here
is the cup''), but as \textsc{LookAt(human,readout)} if the dialogue
context expects the human to look at the readout, and as no event if
no particular look is expected.

The current architecture has an important limitation: The robot
has control of the conversation and directs what is discussed.  This
format is required because of the unreliability of current
off-the-shelf speech recognition tools.  User turns are limited to a
few types of simple utterances, such as ``hello, goodbye, yes, no,
okay,'' and ``please repeat''.  While people often say more complex
utterances,\footnote{In our experimental studies, despite being
  told to limit their utterances to ones similar to those above, some users
  spoke more complex utterances during their conversations with the
  robot.} such utterances cannot be interpreted with any reliability
by current commercially available speech engines unless users train
the speech engine for their own voices.  However, our robot is
intended for all users without any type of pre-training, and therefore
speech and conversation control have been limited.  Future
improvements in speech recognition systems will eventually permit
users to speak complex utterances in which they can express their
desires, goals, dissatisfactions and observations during
collaborations with the robot.  The existing \CollagenTM{} system can
already interpret the intentions conveyed in more complex utterances,
even though no such utterances can be expressed reliably to the robot
at the present time.

Finally, it must be noted here that the behaviors that are supported
in Mel are not found in many other systems.  The MACK screen-based
embodied conversation agent, which uses earlier versions of the same
vision technology used in this work, is also able to point at objects
and to track the human user's head (\cite{nakano03:_face_groun}).
However, the MACK system was tested with just a few users and does not
use the large amount of data we have collected (over more than a year)
of users interacting and nodding to the robot.  This data collection
was necessary to make the vision nodding algorithms reliable enough to
use in a large user study, which we are currently undertaking (see
\cite{morency05:_nodding} for initial results on that work).  A full
report on our experiences with a robot interpreting nodding must be
delayed for a future paper.

\section{Studies with users}
\label{userstudies}

A study of the effects of engagement gestures by the robot with human
collaboration partners was conducted (see \cite{sidner04:_where_look}). The
study consisted of two groups of users interacting with the robot to
collaboratively perform a demo of IGlassware, in a conversation similar to that
described in Figure~\ref{fig:conversation}.  We present the study and
main results as well as additional results related to nodding.  We discuss
measures used in that study as well as additional measures that should
be useful in gauging the naturalness of robotic interactions during
conversations with human users.

Thirty-seven participants were tested across two different conditions.
Participants were chosen from summer staff at a computer science
research laboratory, and individuals living in the local community who
responded to advertisements placed in the community.  Three participants
had interacted with a robot previously; none had interacted with our
robot. Participants ranged in age from 20 to roughly 50 years of age;
23 were male and 14 were female.  All participants were paid a small
fee for their participation.

In the first, the \emph{mover} condition, with 20 participants, the
fully functional robot conducted the demonstration of the IGlassware
table, complete with all its gestures.  In the second, the
\emph{talker} condition, with 17 participants, the robot gave the same
demonstration in terms of verbal utterances, that is, all its
conversational verbal behavior using the speech and \CollagenTM{} system
remained the same.  It also used its visual system to observe the
user, as in the mover condition.  However, the robot was constrained
to talk by moving only its beak in synchrony with the words it spoke.
It initially located the participant with its vision system, oriented
its head to face the user, but thereafter its head remained pointed in
that direction.  It performed no wing or head movements thereafter,
neither to track the user, point and look at objects nor to perform
beat gestures.

In the protocol for the study, each participant was randomly
pre-assigned into one of the two conditions.  Twenty people
participated in the mover condition and 17 in the talker condition.  A
video camera was turned on before the participant arrived.  The
participant was introduced to the robot as ``Mel'' and told the stated
purpose of the interaction, that is, to see a demo from Mel.
Participants were told that they would be asked a series of questions
at the completion of the interaction.

Then the robot was turned on, and the participant was instructed to
approach the robot.  The interaction began, and the experimenter left
the room.  After the demonstration, participants were given a short
questionnaire that contained the scales described in the
Questionnaires section below.  Lastly they also reviewed the videotape
with the experimenter to discuss problems they encountered.

All participants completed the demo with the robot.  Their sessions
were videotaped and followed by a questionnaire and informal
debriefing.  The videotaped sessions were analyzed to determine what
types of behaviors occurred in the two conditions and what behaviors
provided evidence that the robot's engagement behavior approached
human-human behavior.

While our work is highly exploratory, we predicted that people would
prefer interactions with a robot with gestures (the mover condition).
We also expected that participants in the mover condition would
exhibit more interest in the robot during the interaction.  However,
we did not know exactly what form the differences would take.  As our
results show, our predictions are partially correct.

\subsection{Questionnaires}
Questionnaire data focused on the robot's likability, understanding of
the demonstration, reliability/dependability, appropriateness of movement and
emotional response.  

Participants were provided with a post-interaction questionnaire.
Questionnaires were devoted to five different factors concerning the
robot:
\begin{enumerate}
\item \emph{General liking of Mel} (devised for experiment; 3 items).
  This measure
  gives the participants' overall impressions of the robot and their
  interactions with it.
\item \emph{Knowledge and confidence of knowledge of demo} (devised
  for experiment; 6 items).  Knowledge of the demonstration concerns
  task differences.  It was unlikely that there would be a difference
  among participants, but such a difference would be very telling
  about the two conditions of interaction.  Confidence in the
  knowledge of the demonstration is a finer-grained measure of task
  differences.  Confidence questions asked the participant how certain
  they were about their responses to the factual knowledge questions.
  There could potentially be differences in this measure not seen in
  the direct questions about task knowledge.
\item \emph{Involvement in the interaction}
  (adapted from \cite{lombard00:_presense,lombard97:_presence}; 5 items).
  Lombard and Ditton's notion of engagement (different from ours) is a
  good measure of how involving the experience seemed to
  the person interacting with the robot.
\item \emph{Reliability of the robot}
  (adapted from \cite{kidd03:_sociab_robot}, 4 items).
  While not directly related to the outcome of this
  interaction, the perceived reliability of the robot is a good
  indicator of how much the participants would be likely to depend on
  the robot for information on an ongoing basis.  A higher rating of
  reliability means that the robot will be perceived more positively in
  future interactions.
\item \emph{Effectiveness of movements}
  (devised for experiment; 5 items).
  This measure is used to determine the quality of the gestures and
  looking.
\end{enumerate}

Results from these questions are presented in
Table~\ref{tab:questionnaire-results}.  A multivariate analysis of
condition, gender, and condition crossed with gender (for interaction
effects) was undertaken.  No difference was found between the two
groups on likability, or understanding of the demonstration, while a gender
difference for women was found on involvement response.  Participants
in the mover condition scored the robot more often as making
appropriate gestures (significant with $F[1,37] = 6.86$, $p=0.013$, $p
< 0.05$), while participants in the talker condition scored the robot
more often as dependable/reliable ($F[1,37] = 13.77$, $p < 0.001$,
high significance).

\begin{table}
  \centering
  \begin{tabular}[l]{|l|l|}\hline
    \multicolumn{1}{|c|}{\emph{Tested factor}} & \multicolumn{1}{c|}{\emph{Significant effects}}\\\hline\hline
    Liking of Robot:         & \multicolumn{1}{c|}{No effects}\\\hline
    Knowledge of the demo: & \multicolumn{1}{c|}{No effects}\\\hline
    Confidence of knowledge of the demo: & \multicolumn{1}{c|}{No effects}\\\hline
    Engagement in the interaction: & 
    \begin{InTableII}
      \multicolumn{2}{l}{\emph{Effect for female gender:}}\\
      Female average: & 4.84\\
      Male average:   & 4.48\\
      \multicolumn{2}{l}{$F[1,30] = 3.94$}\\
      \multicolumn{2}{l}{$p = 0.0574$ (\emph{Borderline significance})}
    \end{InTableII}\\\hline
    Reliability of robot: &
    \begin{InTableII}
      \multicolumn{2}{l}{\emph{Effect for talker condition:}}\\
      Mover average:  & 3.84\\
      Talker average: & 5.19\\
      \multicolumn{2}{l}{$F[1,37] = 13.77$}\\
      \multicolumn{2}{l}{$p < 0.001$ (\emph{High significance})}\\
    \end{InTableII}\\\hline
    Appropriateness of movements: &
    \begin{InTableII}
      \multicolumn{2}{l}{\emph{Effect for mover condition:}}\\
      Mover average:  & 4.99\\
      Talker average: & 4.27\\
      \multicolumn{2}{l}{$F[1,37] = 6.86$}\\
      \multicolumn{2}{l}{$p=0.013$ ($p < 0.05$: \emph{Significance})}\\\hline
    \end{InTableII}\\\hline
  \end{tabular}
  \caption{Summary of questionnaire results}
  \label{tab:questionnaire-results}
\end{table}
For factors where there are no difference in effects, it is evident
that all participants understood the demonstration and were confident of their
response.  Knowledge was a right/wrong encoding of the answers to the
questions.  In general, most participants got the answers correct (overall
average = 0.94; movers = 0.90; talkers = 0.98).  Confidence was scored
on a 7-point Likert scale.  Both conditions rated highly (overall
average = 6.14; movers = 6.17; talkers = 6.10).  All participants also
liked Mel more than they disliked him.  On a 7-point Likert scale, the
overall average was 4.86.  The average for the mover condition was
4.78, while the talker condition was actually higher, at 4.96.  If one
participant who had difficulty with the interaction is removed, the mover group average becomes 4.88.  None of the comparative differences between
participants is significant.

The three factors with effects for the two conditions provide some
insight into the interaction with Mel.  First consider the effects of
gender on involvement.  The sense of involvement (called engagement in
Lombard and Ditton's work) concerns being ``captured'' by the experience.
Questions for this factor included:

\begin{itemize}
\item  How engaging was the interaction?
\item  How relaxing or exciting was the experience?
\item  How completely were your senses engaged?
\item  The experience caused real feelings and emotions for me.
\item  I was so involved in the interaction that I lost track of time.
\end{itemize}

While these results are certainly interesting, we only conclude that
male and female users may interact in different ways with robots that
fully move.  This result mirrors work by
\cite{shinozawa03:_robot_new_media} who found difference in gender,
not for involvement, but for likability and credibility.
\cite{kidd03:_sociab_robot} found gender differences about how
reliable a robot was (as opposed to an on-screen agent); women found
the robot more reliable, while men found the on-screen agent more so.

Concerning appropriateness of movements, mover participants perceived
the robot as moving appropriately.  In contrast, talkers felt Mel did
not move appropriately.  However, some talker participants said that
they thought the robot moved!  This effect confirms our sense that a
talking head is not doing everything that a robot should be doing in
an interaction, when people and objects are present.  Mover
participants' responses indicated that they thought:

\begin{itemize}
\item  The interaction with Mel was just like interacting with a real person.
\item  Mel always looked at me at the appropriate times.
\item  Mel did not confuse me with where and when he moved his head and wings.
\item  Mel always looked at me when he was talking to me.
\item  Mel always looked at the table and the glass at the appropriate times.
\end{itemize}

However, it is striking that users in the talker  condition found
the robot more reliable when it was just a talking head:
\begin{itemize}
\item  I could depend on Mel to work correctly every time.
\item  Mel seems reliable.
\item  If I did the same task with Mel again, he would do it the same way.
\item  I could trust Mel to work whenever I need him to.
\end{itemize}
There are two possible conclusions to be drawn about reliability: (1)
the robot's behaviors were not correctly produced in the mover
condition, and/or (2) devices such as robots with moving parts are
seen as more complicated, more likely to break and hence less
reliable.  Clearly, much more remains to be done before users are
perfectly comfortable with a robot.

\subsection{Behavioral observations}

What users say about their experience is only one means of determining
interaction behavior, so the videotaped sessions were reviewed and
transcribed for a number of features.  With relatively little work in
this area (see \cite{nakano03:_face_groun} for one study on related
matters with a screen-based ECA), the choices were guided by measures that
indicated interest and attention in the interaction.  These measures
were: 
\begin{itemize}
\item length of interaction time as a measure of overall interest, 
the
\item amount of shared looking (i.e., the combination of time spent
  looking at each other and looking together at objects), as a measure
  of how coordinated the two conversants were,
\item mutual gaze (looking at each other only) also as a measure of
  conversants' coordination,
\item the amount of looking at the robot during the human's turn, as a measure of attention to the robot,
\item  and the amount of looking at the robot overall, also as an attentional measure.
\end{itemize}

\begin{table}
  \centering
  \begin{tabular}[c]{|p{2.5cm}|c|c|l|p{1.9cm}|}\hline
    Measure    &  Mover     & Talker    & Test/Result  &   Significance
    \\\hline\hline
    Interaction time 
    & 217.7 sec
    & 183.1 sec
    & \begin{InTable}Single factor\\ANOVA:\\$F[1,36]=10.34$\end{InTable}
    & Significant: $p < 0.01$ 
    \\\hline
    Shared looking
    & 51.1\%
    & 36.1\%
    & \begin{InTable}Single factor\\ANOVA:\\$F[1,36]= 8.34$\end{InTable}
    & Significant: $p < 0.01$
    \\\hline
    Mutual gaze
    & 40.6\%
    & 36.1\%
    & \begin{InTable}Single-factor\\ANOVA:\\$F[1,36] = 0.74$\end{InTable}
    & No:\break $p = 0.40$
    \\\hline
    \raggedright Speech directed to robot
    & 70.4\%
    & 73.1\%
    & \begin{InTable}Single-factor\\ANOVA:\\$F[1,36]= 4.13$\end{InTable}
    & No:\break $p=0.71$
    \\\hline
    \raggedright Look backs, overall
    & \begin{InTable} 19.65 avg.\\median:\\18-19\end{InTable}
    & \begin{InTable} 12.82 avg.\\median:\\12\end{InTable}
    & \begin{InTable}Single-factor\\ANOVA:\\$F[1,36]=15.00$\end{InTable}
    & Highly:\break $p < 0.001$
    \\\hline
    Table-look 1
    & \begin{InTable}12/19\\(63\%)\end{InTable}
    & \begin{InTable}6/16\\(37.5\%)\end{InTable}
    & \begin{InTable}t-tests\\$t(33)=1.52$\end{InTable}
    & Weak:\break One-tailed: $p=0.07$
    \\\hline
    Table-look 2
    & \begin{InTable}11/20\\(55\%)\end{InTable}
    & \begin{InTable}9/16\\(56\%)\end{InTable}
    & \begin{InTable}t-tests\\$t(34)=-1.23$\end{InTable}
    & No:\break One-tailed:\break $p = 0.47$
    \\\hline
  \end{tabular}
  \caption{Summary of behavior test results in human-robot interaction experiment.}
  \label{tab:behavior-results}
\end{table}

Table~\ref{tab:behavior-results} summarizes the results for the two
conditions.  First, total interaction time in the two conditions
varied significantly (row 1 in Table~\ref{tab:behavior-results}).
This difference may help explain the subjective sense gathered during
video viewing that the talker participants were less interested in the
robot and more interested in doing the demonstration, and hence completed the
interaction more quickly.

While shared looking (row 2 in Table~\ref{tab:behavior-results}) was
significantly greater among mover participants, this outcome is
explained by the fact that the robot in the talker condition could
never look with the human at objects in the environment.  However, it
is noteworthy that in the mover condition, the human and robot spent
51\% of their time (across all participants) coordinated on looking at
each other and the demonstration objects.  Mutual gaze (row 3 in
Table~\ref{tab:behavior-results}) between the robot and human was not
significantly different in the two conditions.

We chose two measures for how humans attended to the robot: speech
directed to the robot during the human's turn, and other times the
human looked back to the robot during the robot's turn.  In the social
psychology literature, \cite{argyle75:_bodily} notes that listeners
generally looked toward the speaker as a form of feedback that they
are following the conversation (p. 162-4). So humans looking at the
robot during the robot's turn would indicate that they are behaving in
a natural conversational manner.

The measure of speech directed to the robot during the human's turn
(row 4 in Table~\ref{tab:behavior-results}) is an average across all
participants as a percentage of the total number of turns per
participant.  There is no difference in the rates.  What is surprising
is that both groups of participants directed their gaze to the robot
for 70\% or more of their turns.  This result suggests that a
conversational partner, at least one that is reasonably sophisticated
in conversing, is a compelling partner, even with little gesture
ability\@.\footnote{We did not eliminate beak movements in the talker
  condition since pre-testing indicated that users found the resulting
  robot non-conversational.}  However, the second measure, the number
of times the human looked back at the robot, are highly significantly
greater in the mover condition.  Since participants spend a good
proportion of their time looking at the table and its objects (55\%
for movers, 62\% for talkers), the fact that they interrupt their
table looking to look back to the robot is an indication of how
engaged they are with it compared with the demonstration objects. This
result indicates that a gesturing robot is a partner worthy of closer
attention during the interaction.

We also found grounding effects in the interaction that we had not
expected.  Participants in both conditions nodded at the robot, even
though during this study, the robot was not able to interpret nods in
any way.  Eleven out of twenty participants in the mover condition
nodded at the robot three or more times during the interaction (55\%)
while in the talker condition, seven out of seventeen participants
(41\%) did.  Nods were counted only when they were clearly evident,
even though participants produced slight nods even more frequently.
The vast majority of these nods accompany ``okay,'' or ``yes,'' while
a few accompany a ``goodbye.''  There is personal variation in nodding
as well.  One participant, who nodded far more frequently than all the
other participants (a total of 17 times), nodded in what appeared to
be an expression of agreement to many of the robot's utterances.  The
prevalence of nodding, even with no evidence that it is understood,
indicates just how automatic this conversational behavior is.  It
suggests that the conversation was enough like a human-to-human
conversation to produce this grounding effect even without planning
for this type of behavior.  The frequency of nodding in these
experiments motivated in part the inclusion of nod understanding in the
robot's more recent behavior repertoire (\cite{lee04:_nodding}).

We also wanted to understand the effects of utterances where the robot
turned to the demonstration table as a deictic gesture.  For the two
utterances where the robot turned to the table (Table-look 1 and 2),
we coded when participants turned in terms of the words in the
utterance and the robot's movements.  These utterances were: ``Right
there \Action{robot gesture} is the IGlassware cup and near it is the
table readout,'' and ``The \Action{robot gesture} copper in the glass
transmits to the readout display by inductance with the surface of the
table.''  For both of these utterances, the mover robot typically (but
not always) turned its head towards and down to the table as its means
of pointing at the objects.  The time in the utterance when pointing
occurred is marked with the label \Action{robot gesture}.  Note that
the talker robot never produced such gestures.

For the first instance, Table-look 1, (``Right there\ldots''), 12/19
mover participants (63\%) turned their heads or their eye gaze during
the phrase ``IGlassware cup.''  For these participants, this change
was just after the robot has turned its head to the table.  The
remaining participants were either already looking at the table (4
participants), turned before it did (2 participants) or did not turn
to the table at all (1 participant); 1 participant was off-screen and
hence not codeable.  In contrast, among the talker participants, only
6/16 participants turned their head or gaze during ``IGlassware cup''
(37.5\%).  The remaining participants were either already looking at
the table before the robot spoke (7 participants) or looked much later
during the robot's utterances (3 participants); 1 participant was off
camera and hence not codeable.

For Table-look 2, (``The copper in the glass\ldots''), 11 mover
participants turned during the phrases ``in the glass transmits,'' 7
of the participants at ``glass.'' In all cases these changes in
looking followed just after the robot's change in looking.  The
remaining mover participants were either already looking at the table
at the utterance start (3 participants), looked during the phrase
``glass'' but before the robot turned (1 participant), or looked
during ``copper'' when the robot had turned much earlier in the
conversation (1 participant). Four participants did not hear the
utterance because they had taken a different path through the
interaction.  By comparison, 12 of the talker participants turned
during the utterance, but their distribution is wider: 9 turned
between ``copper in the glass transmits'' while 3 participants turned
much later in the utterances of the turn. Among the remaining talker
participants, 2 were already looking when the utterance began, 1
participant was distracted by an outside intervention (and not
counted), and 2 participants took a different path through the
interaction.

The results for these two utterances are too sparse to provide strong
evidence. However, they indicate that participants pay attention to when
the robot turns his head, and hence his attention, to the table.  When
the robot does not move, participants turn their attention based on
other factors (which appear to include the robot's spoken utterance,
and their interest in the demonstration table).
\cite{kendon90:_conduction} discusses how human participants in
one-on-one and in small groups follow the head changes of others in
conversation.  Thus there is evidence that participants in this study
are behaving in a way that conforms to their normal human interactions
patterns.

While the results of this experiment indicate that talking encourages
people to respond to a robot, it appears that gestures encourage
them even more.  One might argue that movement alone explains why
people looked more often at the robot, but the talking-only robot does
have some movement---its beak moves.  So it would seem that other gestures
are the critical matter.  The gestures used in the experiment are ones
appropriate to conversation.  It is possible that it is the gestures
themselves, and not their appropriateness in the context of the
conversation, that are the source of this behavior.  Our current
experiment does not allow us to distinguish between appropriate
gestures and inappropriate ones.  However, if the robot were to move
in ways that were inappropriate to the conversation, and if human
partners ignored the robot in that case, then we would have stronger
evidence for engagement gestures.  We have recently completed a set of
experiments that were not intended to judge these effects, but have
produced a number of inappropriate gestures for extended parts of an
interaction.  These results may tell us more about the importance of 
appropriate gestures during conversation.

Developing quantitative observational measures of the effects of
gesture on human-robot interaction continues to be a challenging
problem.  The measures used in this work, interaction time, shared
looking, mutual gaze, looks during human turn, looks back overall,
number of times nodding occurred and in relation to what conversation
events, and observations of the effects of deictic gestures, are all
relevant to judging attention and connection between the human and the
robot in conversation.  The measures all reflect patterns of behavior
that occur in human-human conversation.  This work has assumed that it
is reasonable to expect to find these same behaviors occurring in
human-robot conversation, as indeed they do.  However, there is need
for finer-grained measures, that would allow us to judge more about
the robot's gestures as natural or relevant at a particular point in
the conversation.  Such measures await further research.

\section{Related Research}

While other researchers in robotics are exploring aspects of gesture
(for example \cite{breazeal01:_affect} and
\cite{ishiguro03:_robovie}), none of them have attempted to model
human-robot interaction to the degree that involves the numerous
aspects of engagement and collaborative conversation that we have set
out above.  A robot developed at Carnegie Mellon University serves as
a museum guide (\cite{burgard98:_museum_robot}) and navigates well
while avoiding humans, but interacts with users via a screen-based
talking head with minimal engagement abilities.  Robotics researchers
interested in collaboration and dialogue (e.g., \cite{fong01:_collab})
have not based their work on extensive theoretical research on
collaboration and conversation.  Research on human-robot gesture
similarity (\cite{ono01:_model_gestur_robot}) indicates that body
gestures corresponding to a joint point of view in direction-giving
affect the outcome of human gestures as well as human understanding of
directions.

Our work is also not focused on emotive interactions, in contrast to
Breazeal \cite{breazeal01:_affect} among others (e.g.,
\cite{lisetti04:_social_infor}).

Most similar in spirit to the work reported here is the \textsc{Armar II}
robot (\cite{dillman04:_armar_ii}).
\textsc{Armar II} is speech enabled, has some dialogue capabilities,
and has abilities to track gestures and people.  However, the
\textsc{Armar II} work is focused on teaching the robot new tasks
(with programming by demonstration techniques), while our work has
been focused on improving the interaction capabilities needed to hold
conversations and undertake tasks.  Recently,
\cite{breazeal04:_robot_collab} have explored teaching a robot a
physical task that can be performed collaboratively once learned.

Research on infant robots with the ability to learn mutual gaze and
joint attention (\cite{kozima03:_atten_coupl,nagai03}) offers exciting
possibilities for eventual use in more sophisticated conversational
interactions.

\section{Future work}

Future work will improve the robot's conversational language
generation so that nodding by humans will be elicited more easily.  In
particular, there is evidence in the linguistic literature,
\emph{inter alia} (\cite{clark96:_using_languag}), that human
speech tends to short intonational phrases with pauses for
backchannels rather than long full utterances that resemble sentences
in written text.  By producing utterances of the short variety, we
expect that people will nod more naturally at the robot.  We plan to
test our hypothesis by comparing encounters with our robot where
participants are exposed to different kinds of utterances to test how
they nod in response.

The initiation of an interaction is an important engagement function.
Explorations are needed to determine the combinations of verbal and
non-verbal signals that are used to initially engage a human user in
an interaction (see
\cite{miyauchi04:_activ_eye_contac_human_robot_commun}).  Our efforts
will include providing mobility to our robot as well as extending the
use of current vision algorithms to ``catch the eye'' of the human
user and present verbal feedback in the initiation of engagement.

Current limits on the robot's vision make it impossible to determine
the identity of the user.  Thus if the user leaves and is immediately
replaced by another person, the robot cannot tell that this change has
happened.  Identity recognition algorithms, in variable light without
color features, will soon be used, so that the robot will be able to
recognize the premature end of an interaction when a user leaves.
This capability will also allow the robot to judge when the user might
desire to disengage due to looks away from either the robot or the
objects relevant to collaboration tasks.

Finally, we would like to understand how users change and adapt to the
robot.  Because most of our users have not interacted with robots before, the
novelty of Mel plays some role in their behavior that we cannot
quantify.  We are working on giving the robot several additional
conversational topics, so that users can have several conversations
with Mel over time, and we can study whether and how their behaviors
change.

\section{Conclusions}
In this paper we have explored the concept of engagement, the process
by which individuals in an interaction start, maintain and end their
perceived connection to one another.  We have reported on one aspect
of engagement among human interactors---the effects of tracking faces
during an interaction.  We have reported on a humanoid robot that
participates in conversational, collaborative interactions with
engagement gestures.  The robot demonstrates tracking its human
partner's face, participating in a collaborative demonstration of an
invention, and making engagement decisions about its own behavior as
well as the human's during instances where face tracking was
discontinued in order to track objects for the task.  We also reported
on our findings of the effects on human participants of a robot that
did and did not perform engagement gestures.

While this work is only a first step in understanding the
engagement process, it demonstrates that engagement gestures have an
effect on the behavior of human interactors with robots that converse
and collaborate.  Simply said, people direct their attention to the
robot more often in interactions where gestures are present, and they
find these interactions more appropriate than when gestures are absent.
We believe that as the engagement gestural abilities of robots become
more sophisticated, human-robot interaction will become smoother, be
perceived as more reliable, and will make it possible to include robots
into the everyday lives of people.

\section*{Acknowledgements}

We would like to thank the anonymous reviewers for their thoughtful
 comments and suggestions on this paper.

\bibliography{humanoids}


\end{document}